%% file: main.tex
\documentclass[10pt,twocolumn,letterpaper]{article}

\usepackage[pagenumbers]{cvpr}      
\usepackage{algorithm}      
\usepackage{algpseudocode}  
\usepackage{listings}          
\usepackage{xcolor}            
\usepackage{xcolor}

\usepackage{mwe}
\usepackage{booktabs} 
\usepackage{multirow} 
\usepackage{makecell} 
\usepackage{graphicx} 

\usepackage{fontawesome5}
\newcommand{\method}{\texttt{Yume1.5}\xspace}

\input{preamble}
\definecolor{cvprblue}{rgb}{0.21,0.49,0.74}
\usepackage[pagebackref,breaklinks,colorlinks,allcolors=cvprblue]{hyperref}

\title{\method: A Text-Controlled Interactive World Generation Model}

\author{
    \textbf{Xiaofeng Mao$^{1,2}$, Zhen Li$^{1}$, Chuanhao Li$^{1}$, Xiaojie Xu$^{1}$, Kaining Ying$^{2}$,} \\
    \textbf{Tong He$^{1}$, Jiangmiao Pang$^{1}$, Yu Qiao$^{1}$, Kaipeng Zhang$^{1,3\dagger\ddagger}$}
\\
$^1$\normalfont{Shanghai AI Laboratory}~~~$^2$\normalfont{Fudan University}~~~$^3$\normalfont{Shanghai Innovation Institute}\\
\hspace{0.1em} ~\faGithub~ \hspace{0.4em} Github:\url{https://github.com/stdstu12/YUME}\\
\hspace{0.25em}~\faYoutube~  \hspace{0.4em} Project Page: \url{https://stdstu12.github.io/YUME-Project}\\
\hspace{0.35em}~\faDatabase~ \hspace{0.55em} Data: \url{https://github.com/Lixsp11/sekai-codebase}
}

\makeatletter
\newcommand{\thickhline}{%
    \noalign {\ifnum 0=`}\fi \hrule height 1pt
    \futurelet \reserved@a \@xhline
}

\makeatletter

\begin{document}






\twocolumn[{
\renewcommand\twocolumn[1][]{#1}%
\maketitle
\begin{center}
    \centering
    \captionsetup{type=figure}
    \includegraphics[width=1\linewidth]{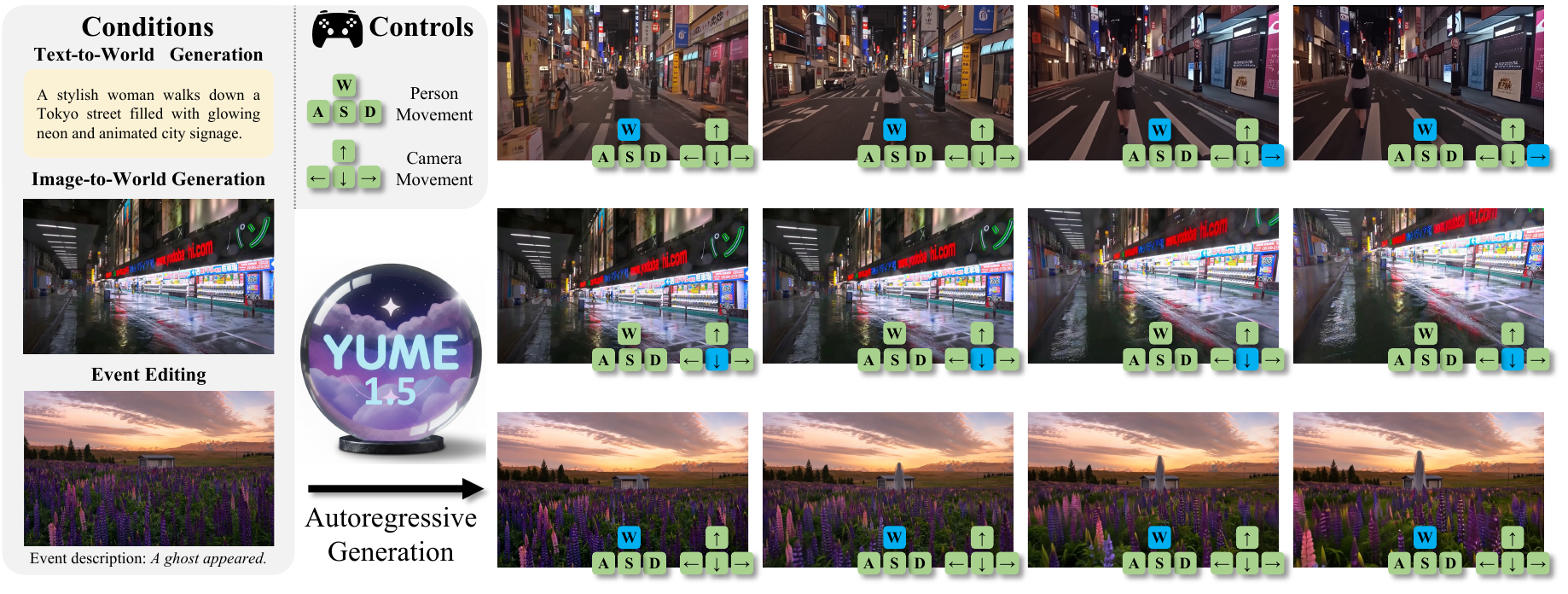}
    \vspace{-6mm}
    \captionof{figure}{Our \textbf{\method} framework supports three interactive generation modes: text-to-world generation from descriptions, image-to-world generation from static images, and text-based event editing. All modes are controlled through continuous keyboard inputs for person and camera movements, enabling autoregressive generation of explorable and persistent virtual worlds. We have included \textbf{demo} videos in the supplementary materials.}
\vspace{3mm}
\label{fig:teaser}
\end{center}
}]

\input{sec/0_abstract}    
\input{sec/1_intro} 
\input{sec/2_rele} 
\input{sec/3_method} 
\input{sec/4_exprment}
{
    \small
    \bibliographystyle{ieeenat_fullname}
    \bibliography{main}
}


\end{document}

%% file: sec/0_abstract.tex
\if TT\insert\footins{\noindent\footnotesize\\
    $^{\dagger}$~Project Leader\\
    $^{\ddagger}$~Corresponding Author \\
}\fi

\begin{abstract}
Recent approaches have demonstrated the promise of using diffusion models to generate interactive and explorable worlds. However, most of these methods face critical challenges such as excessively large parameter sizes, reliance on lengthy inference steps, and rapidly growing historical context, which severely limit real-time performance and lack text-controlled generation capabilities.
To address these challenges, we propose \method, a novel framework designed to generate realistic, interactive, and continuous worlds from a single image or text prompt. \method achieves this through a carefully designed framework that supports keyboard-based exploration of the generated worlds. The framework comprises three core components: (1) a long-video generation framework integrating unified context compression with linear attention; (2) a real-time streaming acceleration strategy powered by bidirectional attention distillation and an enhanced text embedding scheme; (3) a text-controlled method for generating world events. We have provided the codebase in the supplementary material. 
The \textbf{model weights} and \textbf{full codebase} will be made public. 
\end{abstract}

%% file: sec/1_intro.tex
\section{Introduction}
\label{sec:introduction}

Video diffusion models~\cite{Singer2023makeavideo,BarTal2024lumiere,Blattmann2023stablevideo,Ma2024hunyuanvideo,Nagrath2024mochi,Singer2023makeavideo,StepVideo2025technical,Chen2025skyreelsv2}, which have shown remarkable capabilities in synthesizing high-fidelity and temporally coherent visual content~\cite{Ho2022imagenvideo, Blattmann2023stablevideo}, present a promising avenue to realize such a sophisticated interactive world generation task. Recently, the automated generation of vast, interactive, and persistent virtual worlds~\cite{zhang2025matrixgame,agarwal2025cosmos} has advanced rapidly, driven by progress in generative models and the growing demand for immersive experiences in domains such as world simulation, interactive entertainment~\cite{li2025hunyuan}, and virtual embodiment~\cite{xiao2025worldmem, zhang2025matrixgame}.

However, existing video diffusion methods face significant challenges in generating interactive and realistic videos. The main bottlenecks include:
\textbf{(1) Limited Generalizability:} Most methods~\cite{zhang2025matrixgame,xiao2025worldmem} are trained on game datasets, creating a domain gap that makes it difficult to generate realistic dynamic urban scenes. \textbf{(2) High Generation Latency:} The high computational cost of diffusion models hinders real-time continuous generation required for unlimited exploration, thus limiting interactivity. \textbf{(3) Insufficient Text Control Capability:} Existing methods~\cite{zhang2025matrixgame,xiao2025worldmem} utilize images to generate videos and only keyboard, mouse  control is supported but lack text control capability and cannot generate random events.

To address these limitations, we propose the \method, which generates interactive infinite video worlds from single images in an autoregressive manner~\cite{mao2025yume}. Through systematic optimization across three key dimensions, \method achieves intuitive and stable camera control via keyboard inputs while significantly enhancing visual quality and continuity in complex scene generation:

\textbf{(1) Long Video Generation via Joint Temporal-Spatial-Channel Modeling~(TSCM).} We observe that existing methods suffer from either slow inference speed~\cite{zhang2025packing} as video duration increases. To overcome this, we design a joint temporal-spatial and channel compression approach: historical frames are compressed along temporal-spatial dimensions for input to the DiT~\cite{dit}, while channel-wise compressed features are processed by a parallel linear DiT. This design significantly reduces memory consumption and improves inference speed.

\textbf{(2) Acceleration Method.} Through synergistic optimization combining score distillation, we substantially enhance sampling efficiency while maintaining visual quality. Considering that the model suffers from error accumulation during inference, where fewer inference steps lead to more severe error accumulation, we designed an approach similar to Self-Forcing~\cite{huang2025self} to mitigate this issue. Unlike Self-Forcing, however, we replace the KV cache with our proposed TSCM, establishing a novel training paradigm.

\textbf{(3) Text-Controlled World Event Generation.} Observing the absence of text-based generation capabilities in prior models~\cite{mao2025yume,zhang2025matrixgame}, we enable event generation through careful architectural design and a mixed-dataset training strategy, achieving this capability with minimal data requirements.

In summary, \method represents a significant advancement in generating high-quality, dynamic, and interactive worlds. Our main contributions are as follows:

\begin{itemize}
    \item We propose Joint Temporal-Spatial-Channel Modeling (TSCM) for infinite-context generation, which maintains stable sampling speed despite increasing context length.
    \item We integrate Self-Forcing with TSCM to accelerate \method's inference while reducing error accumulation.
    \item Through careful dataset and model architecture design, \method achieves superior performance on both world generation and editing.
\end{itemize}

%% file: sec/2_rele.tex
\section{Related Works}
\label{sec:RelatedWorks}

\subsection{World Generation with Video Diffusion Models}

Diffusion models~\cite{SohlDickstein2015deep, Ho2020denoising}, initially for image synthesis, now underpin video generation. The adoption of Latent Diffusion Models (LDMs) was pivotal for efficiency, leading to works like Video LDM~\cite{Blattmann2023align} which integrated temporal awareness for high-resolution video. Early text-to-video breakthroughs, including Imagen Video~\cite{Ho2022imagenvideo} and Make-A-Video~\cite{Singer2023makeavideo}, demonstrated the potential of this domain. The field has since advanced significantly in scale and architecture. Large-scale models like Google's Lumiere~\cite{BarTal2024lumiere}, featuring a Space-Time U-Net, and OpenAI's Sora~\cite{Brooks2024sora}, with its diffusion transformer, have greatly improved the generation of long, coherent, and high-fidelity videos. Concurrently, the open-source ecosystem has flourished, with Stable Video Diffusion~\cite{Blattmann2023stablevideo} providing a robust baseline, and recent efforts such as HunyuanVideo~\cite{Ma2024hunyuanvideo}, MoChi-Diffusion-XL~\cite{Nagrath2024mochi}, Step-Video-T2V~\cite{StepVideo2025technical}, and SkyReels-V2~\cite{Chen2025skyreelsv2} continue to push the boundaries. Building on these powerful video generation capabilities, recent research focuses on creating directly controllable worlds. Examples include Genie~\cite{bruce2024genie} for generating action-controllable 2D worlds, and GAIA-1~\cite{Wayve2023gaia1} for realistic driving simulations. To ensure long-term consistency, which is critical for exploration, frameworks like StreamingT2V~\cite{henschel2024streamingt2v} for extendable videos, Matrix-Game~\cite{zhang2025matrixgame} for interactive games, and WORLDMEM~\cite{xiao2025worldmem} for memory-enhanced coherence have been proposed. Based on these advanced foundation models, we built \method for realistic dynamic world exploration.

\subsection{Camera Control in Video Generation}
\label{sec:related_cam_control}
Precise camera control is crucial for customizable video generation. While early models lacked explicit control mechanisms, recent research has focused on conditioning video diffusion models directly on camera parameters. A significant body of work enables control via explicit camera trajectories. MotionCtrl~\cite{wang2023motionctrl} introduced a unified controller for camera and object motion using pose sequences. Direct-a-Video~\cite{yang2024directavideo} provided decoupled control of camera pan/zoom, and CameraCtrl~\cite{he2024cameractrl} proposed a plug-and-play module for integrating pose control into existing models. This was later extended by CameraCtrlII~\cite{zhang2024cameractrlii} for iterative, long-form video exploration. Training-free approaches have also emerged, such as CamTrol~\cite{geng2024trainingfree}, which manipulates latent noise priors to guide the camera without model fine-tuning. These methods typically rely on sequences of absolute camera poses to define the trajectory, requiring precise specification of camera parameters at each timestep. In contrast to these approaches that require fine-grained adjustments to explicit camera paths, we achieves intuitive keyboard-based control by discretizing the camera pose space.

\begin{figure}[t]
    \centering
    \includegraphics[width=1.0\linewidth]{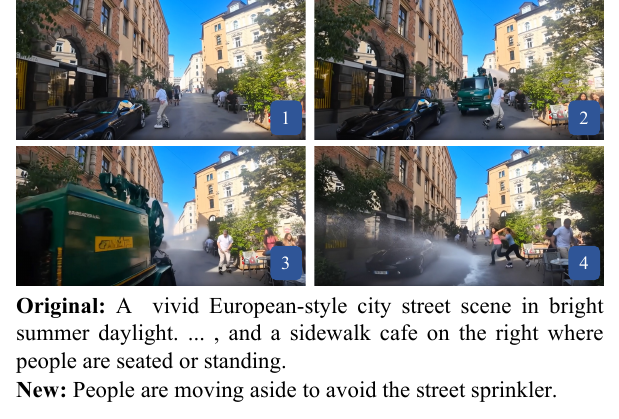}
    \vspace{-5mm}
    \caption{
    An example of re-annotating the dataset. The original and new captions are used for T2V and I2V training, respectively. The \textbf{Original} caption describes detail scene context, while the \textbf{New} caption, generated by VLM, explicitly focuses on dynamic events. 
}
    \label{fig:long_video}
\end{figure}

\subsection{Long Video Generation}

Several recent works have explored methods for long video generation. SkyReels V2~\cite{chen2025skyreels} adopts a diffusion forcing~\cite{chen2024diffusion} framework and employs a sliding window approach, using the last few generated frames as "historical context" or conditioning to predict and generate subsequent video segments. CausVid~\cite{yin2025slow} combines KV cache~\cite{vaswani2017attention} with a diffusion model to enable autoregressive inference. Self-Forcing~\cite{huang2025self} improves upon CausVid by forcing the model to predict the next frame based on its own previously generated frames (which already contain errors), thereby reducing the error accumulation observed in CausVid. However, these methods do not adequately address the issue of increasing memory consumption or computational load as the generated video length grows, and merely truncate previous frames in a simplistic manner. Unlike these methods, our approach does not require prematurely truncating historical context information to alleviate computational load.

\begin{figure*}[t]
    \centering
    \includegraphics[width=1.0\linewidth]{figures/yume2.pdf}
    \vspace{-5mm}
    \caption{Core components of \method. (a) DiT Block with linear attention for efficient feature fusion. (b) Training pipeline with decomposed event and action descriptions. (c) Adaptive history tokens downsampling with varying compression rates based on temporal distance. (d) Chunk-based autoregressive inference with dual-compression memory management.}
    \label{fig:pipeline}
\end{figure*}

%% file: sec/3_method.tex
\section{Data Processing}
\label{sec:Data}

To enhance model training and enable both Text-to-Video (T2V) and Image-to-Video (I2V) capabilities, we constructed a comprehensive dataset by integrating three distinct sources: a Real-world Dataset, a Synthetic Dataset, and a specialized Event Dataset. These components were meticulously processed and combined to balance the model's performance across realistic motion control, general video quality, and specific event generation.

\subsection{Real-world Dataset}
We employ Sekai-Real-HQ as our primary real-world training source. This dataset constitutes a subset of Sekai~\cite{li2025sekai}, containing large-scale walking video clips with high-quality annotations, including camera motion trajectories and semantic labels. To adapt this dataset for our specific requirements, we implemented two key modifications:

First, we utilize the method from \cite{mao2025yume} to derive keyboard and mouse control signals from the trajectory data. These signals are conditioned to guide the model's generation. We map these controls into discrete action descriptions using the vocabularies defined in Eq.~\ref{eq:vocab_camera} and Eq.~\ref{eq:vocab_human}:

\begin{equation}
\label{eq:vocab_camera}
\resizebox{\linewidth}{!}{%
$\mathrm{vocab\_camera} = \left\{
\begin{array}{l}
\rightarrow: \text{Camera turns right ($\rightarrow$).}\\
\leftarrow: \text{Camera turns left ($\leftarrow$)}\\
\uparrow: \text{Camera tilts up ($\uparrow$)}\\
\downarrow: \text{Camera tilts down ($\downarrow$)}\\
\uparrow \rightarrow: \text{Camera tilts up and turns right ($\uparrow\rightarrow$)}\\
\downarrow \rightarrow: \text{Camera tilts down and turns right ($\downarrow\rightarrow$)}\\
\downarrow \leftarrow: \text{Camera tilts down and turns left ($\downarrow\leftarrow$)}\\
\cdot: \text{Camera remains still ($\cdot$)}.
\end{array}
\right.$
}
\end{equation}

\begin{equation}
\label{eq:vocab_human}
\resizebox{\linewidth}{!}{%
$\mathrm{vocab\_human} = \left\{
\begin{array}{l l}
\text{W : Camera moves forward (W).} \\
\text{A : Camera moves left (A).} \\
\text{S : Camera moves backward (S).} \\
\text{D : Camera moves right (D).} \\
\text{W+A : Camera moves forward and left (W+A).} \\
\text{W+D : Camera moves forward and right (W+D).} \\
\text{S+D : Camera moves backward and right (S+D).} \\
\text{S+A : Camera moves backward and left (S+A).} \\
\text{None : Camera stands still ($\cdot$).}
\end{array}
\right.$
}
\end{equation}

Second, we re-annotated the dataset to distinguish between T2V and I2V tasks. While the original annotations—describing scenes and context—are retained for Text-to-Video training, we employ InternVL3-78B~\cite{zhu2025internvl3} to generate new descriptions for Image-to-Video training. These new captions focus specifically on the \textit{events} occurring within the video rather than the static scene context, thereby enhancing the model's event-driven generation capabilities. Figure~\ref{fig:long_video} summarizes the differences between these annotation strategies.

\subsection{Synthetic Dataset}
Since our model is initialized using a pre-trained video diffusion model, relying solely on real-world data may lead to catastrophic forgetting. To mitigate this and avoid overfitting to the Sekai-Real-HQ domain, we incorporate a high-quality synthetic dataset. We start with the Openvid~\cite{nan2024openvid} dataset, performing similarity-based caption deduplication and random sampling to select 80,000 diverse captions. Using Wan 2.1~\cite{wan2025wan} 14B, we synthesized 80,000 videos at 720p resolution. We then computed quality scores using VBench~\cite{huang2024vbench} and filtered the results to retain the top 50,000 videos. These samples are primarily used for text-to-video training to maintain the model's general video generation ability.

\subsection{Event Dataset}
To further bolster the model's capability in generating specific events, we created a specialized Event Dataset. We recruited volunteers (compensated at rates meeting or exceeding local minimum wage standards) to write descriptions across four distinct categories: urban daily life (e.g., cats playing), sci-fi (e.g., UFO encounters), fantasy (e.g., dragons breathing fire), and weather phenomena (e.g., sudden heavy rainfall). 

We collected 10,000 first-person perspective images corresponding to these descriptions and utilized Wan 2.2 14B-I2V\footnote{https://huggingface.co/Wan-AI/Wan2.2-I2V-A14B} to synthesize 10,000 image-to-video sequences. Through rigorous manual screening, we selected 4,000 videos that accurately matched their scene descriptions. These curated videos are employed specifically for text-to-video training to improve semantic alignment in complex scenarios.

\section{Method}
\label{sec:Method}

We propose a comprehensive framework for generating interactive, realistic, and temporally coherent video worlds through systematic innovations across multiple dimensions. Our approach establishes a unified foundation for joint text-to-video and image-to-video generation while addressing key challenges in long-term consistency and real-time performance. The core contributions include: (1) a joint TSCM strategy for efficient long-video generation; (2) a real-time acceleration framework combining TSCM and Self-Forcing; and (3) an alternating training paradigm that enables both world generation and exploration capabilities. Collectively, these advancements facilitate the creation of dynamic, interactive environments suitable for complex real-world scene exploration.

\subsection{Architecture Preliminary}
We establish a foundational model for joint text-to-video and image-to-video generation by adopting the methodology proposed by Wan~\cite{wan2025wan}. This approach initializes the video generation process using a noise: $z \in \mathbb{R}^{C \times f_t \times h \times w}$. 

For text-to-video training, the text embedding $c$ and $z$ which is then fed into the DiT backbone. 

For the image-to-video model, given an image or video condition \(z_c \in \mathbb{R}^{C \times f_i \times h \times w}\), it is zero-padded to match the dimensions \(C \times f_t \times h \times w\). A binary mask \(M_c \in \{0,1\}^{1 \times f_t \times h \times w}\) is constructed (where 1 indicates preserved regions and 0 denotes regions to be generated). The conditional input is fused via \(M_c \cdot z_c + (1 - M_c) \cdot z\) and subsequently processed by the Wan DiT backbone. At this point, $z$ can be considered as consisting of historical frames $z_{c}$ and predicted frames $z_{p}$.

Our text encoding strategy differs from Wan's approach. While Wan processes the entire caption directly through T5, we decompose the caption into Event Description and Action Description as shown in Figure~\ref{fig:pipeline} (b), feeding them separately into T5~\cite{raffel2020exploring}. The resulting embedded representations are concatenated afterward. The Event Description specifies the target scene or event to be generated, while the Action Description defines keyboard and mouse controls. This methodology offers significant advantages: since the set of possible Action Descriptions is finite, they can be precomputed and cached efficiently. Meanwhile, the Event Description is processed only during the initial generation phase. As a result, our approach substantially reduces T5 computational overhead during subsequent video inference steps.
The model is trained using the Rectified Flow loss~\cite{liu2022flow}.

\subsection{Long Video Generation via Joint Temporal-Spatial-Channel Modeling~(TSCM)}

Given the extended duration of video inference, the frame count \(f_i\) of the video condition \(z_c\) progressively increases, leading to substantial computational overhead. It is impractical to include all contextual frames in the computation. Several existing methods aim to mitigate this issue: 

\textbf{(1) Sliding Window:} A widely adopted approach that selects consecutive recent frames within a window near the current prediction frame. This method, however, tends to result in the loss of historical frame information.

\textbf{(2) Historical Frame Compression:} Methods such as FramePack~\cite{zhang2025packing} and Yume~\cite{mao2025yume} compress historical frames, applying less compression to frames closer to the prediction frame and greater compression to those farther away. This similarly leads to increased information loss in more distant historical frames. 

\textbf{(3) Camera Trajectory-Based Search:} Approaches like World Memory~\cite{xiao2025worldmem} leverage known camera trajectories to compute the field-of-view overlap between historical frames and the current frame to be predicted, selecting frames with the highest overlap. This method is incompatible with video models controlled via keyboard input. Even with predicted camera trajectories, it remains difficult to accurately estimate the trajectory under dynamic viewpoint changes, often resulting in significant errors.

To address these limitations, our method proposes a joint Temporal–Spatial–Channel modeling approach, implemented in two steps. We consider applying temporal-spatial compression and channel-wise compression separately to the historical frames \( z_c \).

\subsubsection{Temporal-Spatial Compression}

For historical frames $z_c$, we first apply temporal and spatial compression: we perform random frame sampling at a rate of 1 out of 32, followed by using a high-compression-ratio Patchify. The compression scheme operates as follows:
\begin{equation}
\begin{aligned}
&\text{Frames } t-1 \text{ to } t-2: & (1, 2, 2) \\
&\text{Frames } t-3 \text{ to } t-6: & (1, 4, 4) \\
&\text{Frames } t-7 \text{ to } t-23: & (1, 8, 8) \\
&\vdots & \vdots \\
&\text{Initial frame:} & (1, 2, 2)
\end{aligned}
\end{equation}

\begin{figure*}[htp]
    \centering
    \includegraphics[width=0.9\linewidth]{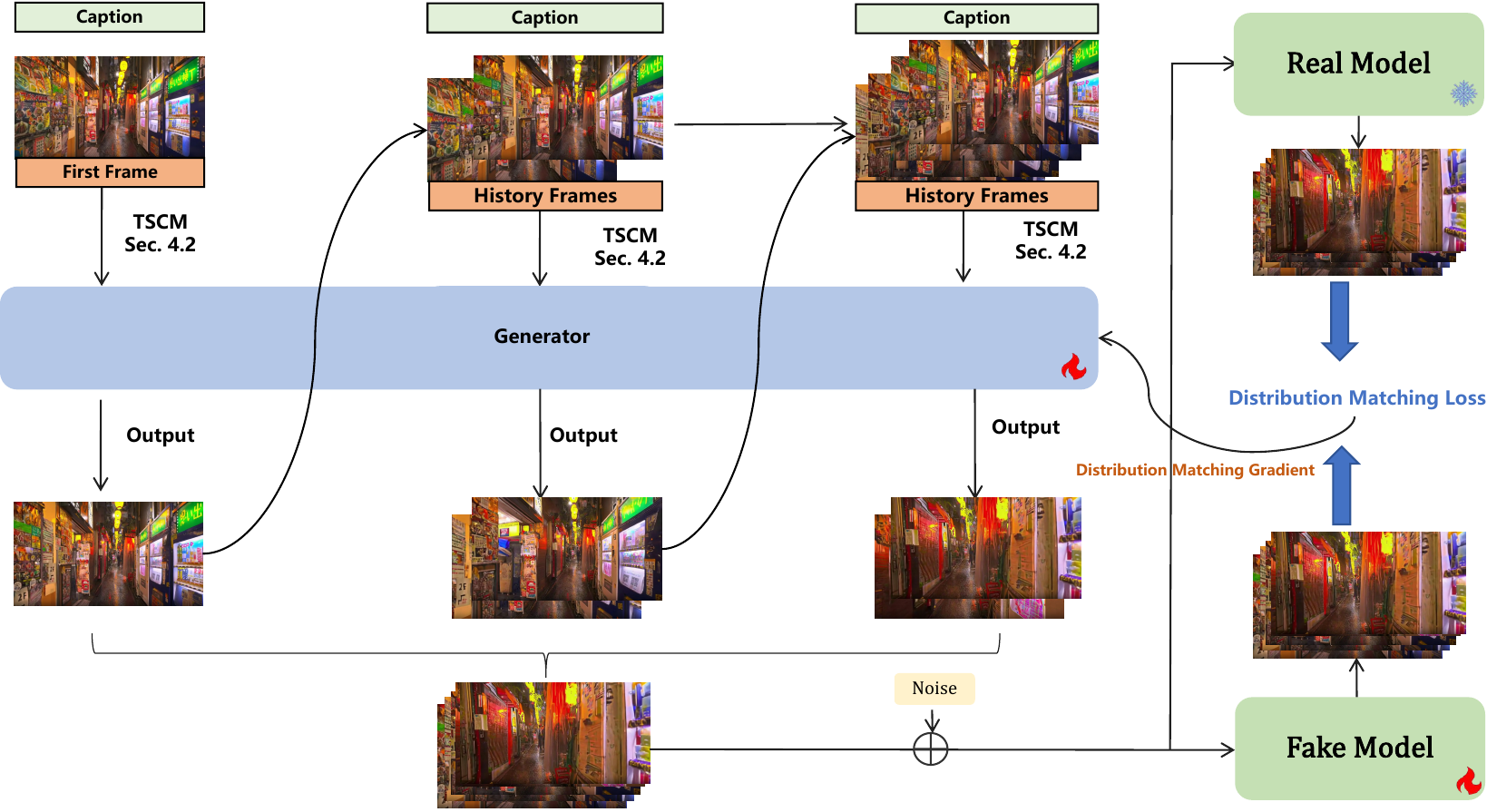}
    \caption{Long-form video generation method. \textbf{Left (Generator):} The model autoregressively generates video chunks. Critically, it uses its \textit{own generated frames} (rather than ground truth) as historical context—compressed by \textbf{TSCM} (Sec. 4.2)—to mitigate the train-inference discrepancy.
  \textbf{Right (Distillation):} The \textbf{Fake Model} (student) is optimized to match the trajectory of the \textbf{Real Model} (teacher) via a distribution matching gradient. This enables high-quality few-step inference while robustly handling error accumulation in long videos.}
    \label{fig:long}
\end{figure*}

Here, (1, 2, 2) denotes downsampling rates of $1\times$, $2\times$, and $2\times$ along the temporal, height, and width dimensions of \(z_c\), respectively. Similarly, (1, 4, 4) corresponds to downsampling rates of $1\times$, $4\times$, and $4\times$ along the same dimensions, and so forth. We achieve these varying downsampling ratios by interpolating the Patchify weights within the DiT. Compared to YUME, our approach of performing temporal random frame sampling reduces both the parameter count of Patchify and the computational load of the model. We obtain the compressed representation \(\hat{z}_c\) and process the prediction frame through the original Patchify with a downsampling rate of (1, 2, 2). The compressed representation \(\hat{z}_c\) is then concatenated with the processed prediction frames \(\hat{z}_p\), and the combined tensor is fed into the DiT Block.

\subsubsection{Channel Compression}
We apply further downsampling to the historical frames \(z_c\), we pass the $z_c$ through a Patchify with a compression rate of (8, 4, 4) and the channel dimension to 96, obtaining \(z_{\text{linear}}\). As shown in Figure~\ref{fig:pipeline} (a), these compressed historical tokens are fed into the DiT block. After the video tokens \(z^l\) pass through the cross-attention layer in the DiT block, they are first processed by a fully connected (FC) layer for channel reduction, after extracting the predicted frames $z^l_p$, concatenate them with \(z_{\text{linear}}\). The combined tokens $z_{fus}$ are fused via a linear attention layer to produce \(z^l_{\text{fus}}\). Finally, \(z^l_{\text{fus}}\) is passed through another FC layer to restore the channel dimension and then added element-wise to \(z_l\) for feature fusion: \(z^l_{\text{fus}}[:, -N_l:] + z_l\), where $N_l$ denotes the number of tokens in the $z_l$.

\noindent \textbf{linear attention.} Our design is illustrated in Figure~\ref{fig:pipeline} (a). This approach draws inspiration from Linear Attention~\cite{shen2021efficient}, which is straightforward to implement. We project \( z^l_{fus} \) through fully connected layers to obtain the query \(q^l\), key \(k^l\), and value \(v^l\) representations, then replace the exponential kernel function \(\exp((k^l)^T q^l)\) with a dot product \(\phi(k^l)^T \phi(q^l)\), where \(\phi: \mathbb{R}^d \to \mathbb{R}^{n}\) is the ReLU activation function. The computation is defined as follows:

\begin{equation}
    o^l = \frac{ \left(\sum_{i=1}^N v^l_i \phi( k^l_i)^T \right)\phi(q^l)}{\left(\sum_{j=1}^{N}\phi(k^l_j)^T\right)\phi(q^l)}
\end{equation}
where \( N \) denotes the number of tokens in \( z^l_{fus} \). We then compute ${\left(\sum_{j=1}^{N}\phi(k^l_j)^T\right)\phi(q^l)}$ prior to applying ROPE to \(q_l\) and \(k_t\), while incorporating normalization layers \(q=\mathrm{Norm}(q), \quad k=\mathrm{Norm}(k)\) to prevent gradient instability. Typically, the attention output \( o_t \) is passed through a linear layer, so we apply a normalization layer before this computation:
\begin{equation}\label{eq29}
\hat{o}_t = \mathrm{Norm}(o_t) W^o
\end{equation}

\noindent \textbf{Summary.} Since the computational cost of standard attention is sensitive to the number of input tokens, we compress historical frames via temporal-spatial compression and process them alongside the prediction frame using standard attention within the DiT block. In contrast, as linear attention is sensitive to the channel dimension, we apply channel-wise compression to historical frames and fuse them with the prediction frame in the linear attention layer of the DiT block. Through this approach, we achieve joint temporal-spatial-channel compression while preserving generation quality.

\subsection{Real-time Accelerate} 
We first train the pre-trained diffusion model on a mixed dataset. We employ an alternating training strategy for text-to-video and image-to-video tasks. Specifically, the model is trained on text-to-video datasets at the current step and switched to image-to-video datasets at the next step. This approach equips the model with comprehensive capabilities in world generation, editing, and exploration. The resulting model is called the foundation model.

As illustrated in Figure~\ref{fig:long}, we first initialize the generator \(G_\theta\), fake model \(G_s\), and real model \(G_t\) with weights from a foundation model~\cite{huang2025self}. The generator samples previous frames from its own distribution and uses them as context to generate new predicted frames. This process iterates, sequentially producing and assembling frames to form a clean video sequence \(z_0\). We then convert the multi-step diffusion model into a few-step generator \(G_\theta\)~\cite{yin2024one} by minimizing the expected KL divergence between the diffused real data distribution and the generated data distribution across noise levels \(t\):
\begin{equation}
\small
\begin{aligned}
s_{\text{real}}(z_t, t) &= \nabla_{z_t} \log p_{\text{real},t}(z_t) \\
&= - \frac{z_t - \alpha_t G_\text{real}(z_t, t)}{\sigma_t^2}, \\
s_{\text{fake}}(z_t, t) &= \nabla_{z_t} \log p_{\text{fake},t}(z_t) \\
&= - \frac{z_t - \alpha_t G_\text{fake}(z_t, t)}{\sigma_t^2}, \\
\nabla\mathcal{L}_\text{DMD} &= - \mathbb{E}_t\left(\int \big(s_{\text{real}}(F(G_{\theta}(z_t), t), t) \right. \\
&\quad \left. - s_{\text{fake}}(F(G_{\theta}(z_t), t), t)\big) \frac{dG_\theta(z)}{d\theta} dz\right).
\end{aligned}
\label{eq:kl-grad}
\end{equation}
where \(F\) is the forward diffusion at step \(t\)~\cite{yin2024improved}. The key distinction from DMD lies in using model-predicted data rather than real data as video conditioning, thereby alleviating the train-inference discrepancy and associated error accumulation.

Our approach diverges from Self-Forcing by eliminating the KV cache and introducing Temporal-Spatial-Channel Modeling (TSCM), enabling utilization of substantially longer contextual information.

%% file: sec/4_exprment.tex
\section{Experiment}
\label{sec:Experiment}

\subsection{Experimental Settings}
\subsubsection{Training Details}
We utilized the Wan2.2-5B\footnote{https://huggingface.co/Wan-AI/Wan2.2-TI2V-5B} as the pre-trained model.
We first conducted foundation model training with the following configuration: video resolution of 704$\times$1280, frame rate of 16 FPS, batch size of 40, and the Adam optimizer with a learning rate of 1e-5. The training was performed on NVIDIA A100 GPUs for 10,000 iterations. Subsequently, we conducted Self-Forcing with TSCM training under identical hyperparameters, except the number of iterations was reduced to 600.
\subsubsection{Evaluation Dataset}
We employ the Yume-Bench~\cite{mao2025yume} evaluation framework. Yume-Bench assesses two core capabilities of the model: visual quality and instruction following (camera motion tracking), using six fine-grained metrics. For instruction tracking evaluation, we examine whether the generated videos correctly adhere to the intended walking direction and camera movements. For the remaining metrics, we adopt the metrics from VBench \cite{huang2024vbench}, including subject consistency, background consistency, motion smoothness, aesthetic quality, and imaging quality. The test data has a resolution of 544 ×960, a frame rate of 16 FPS, and consists of 96 frames in total. We applied 4 inference steps for \method.

\subsection{Quantitative Results}
\subsubsection{Image-to-Video Generation}
We compared several state-of-the-art (SOTA) image-to-video generation models, including Wan-2.1 and MatrixGame, as shown in Table \ref{tab:1}.
Our experimental results revealed that:
(1) Wan-2.1~\cite{wan2025wan} and MatrixGame~\cite{zhang2025matrixgame} shows limited instruction-following capabilities in real world testset. 
(2) Although Yume~\cite{mao2025yume} demonstrates some degree of controllability, it struggles to generalize to real-world scenarios and lacks sufficient scene replication control. In contrast, our \method excels in controllability, with its instruction-following capability scoring 0.836, significantly outperforming other models.
Although Yume~\cite{mao2025yume} demonstrates certain controllability, its reasoning capability remains limited. In comparison to these models, our \method excels in controllability, with its instruction-following score reaching 0.836, significantly outperforming other models. We find that \method achieves an average generation speed of 12 fps at 540p resolution using only a single A100 GPU. We provide Text-to-Video (T2V) VBench metrics in the appendix.
\begin{table}[]
    \small
    \centering
    \renewcommand\arraystretch{1.1}
    \setlength{\tabcolsep}{1.9pt}
  \caption{Quality comparison of different models. IF: Instruction Following, SC: Subject Consistency, BC: Background Consistency, MS: Motion Smoothness, AQ: Aesthetic Quality, IQ: Imaging Quality. Wan-2.1 utilize text-based control. MatrixGame employs its own native keyboard/mouse control scheme. 
  }
  \label{tab:1}
  \vspace{-3mm}
  
  \begin{tabular}{l|c|c|c|c|c|c|c}
    \thickhline
    Model
    & Time(s)$\downarrow$
    & IF$\uparrow$ & SC$\uparrow$ & BC$\uparrow$ & MS$\uparrow$ & AQ$\uparrow$ & IQ$\uparrow$ \\
    \hline
    Wan-2.1    &611 &0.057 &0.859 &0.899 &0.961 &0.494 &0.695\\
    MatrixGame&971 &0.271 &0.911 &0.932 &0.983 &0.435 &\textbf{0.750} \\
    Yume &572&0.657&\textbf{0.932} &\textbf{0.941} &\textbf{0.986} &\textbf{0.518} &0.739 \\
    \method &8&\textbf{0.836}&\textbf{0.932} &\textbf{0.945} &{0.985} &\textbf{0.506} &0.728  \\
    \thickhline
  \end{tabular}
\end{table}

\begin{figure}[htp]
    \centering
    \includegraphics[width=\linewidth]{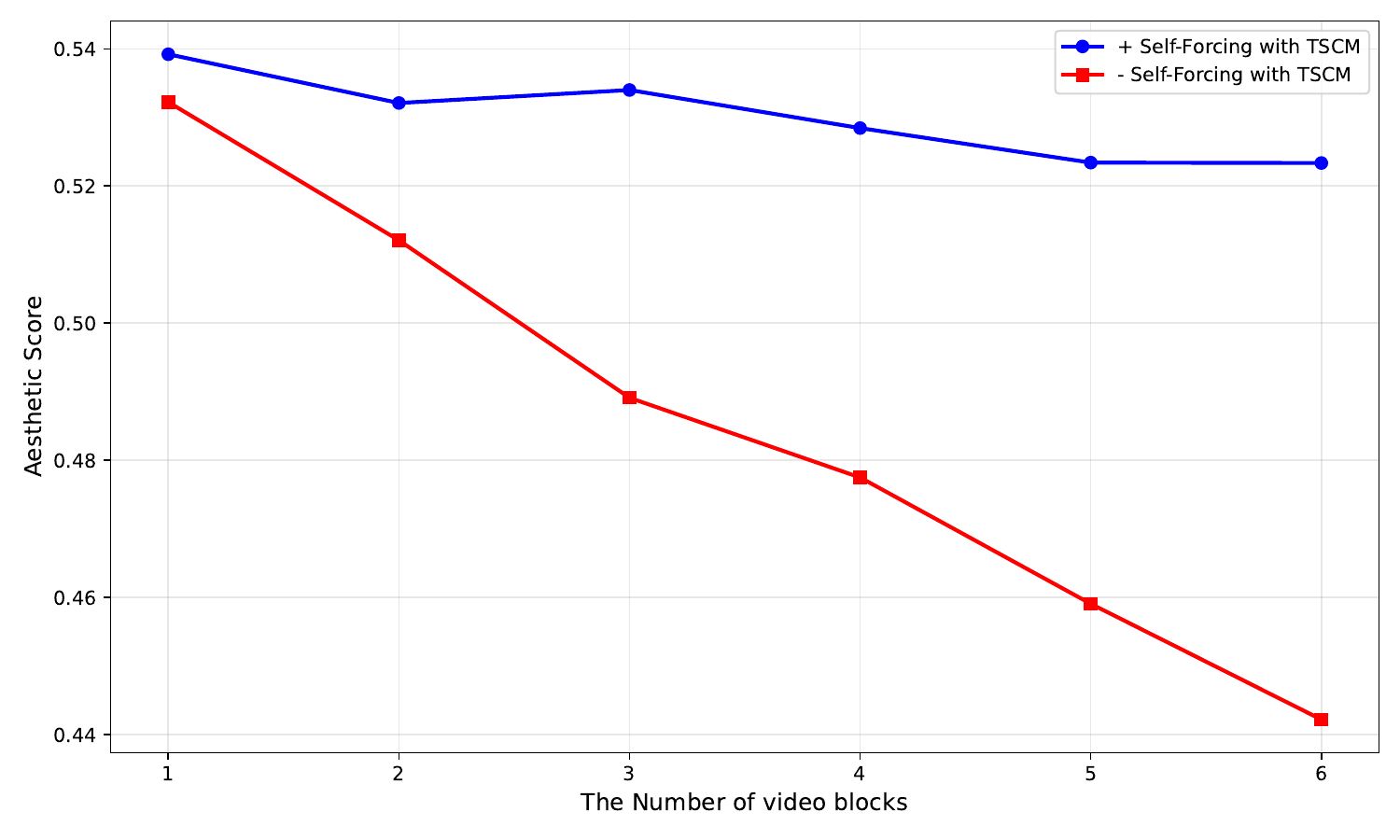}
    \caption{Aesthetic Score Dynamics in Long-video Generation. Aesthetic Score Dynamics in Long-video Generation. The~\textbf{x-axis} represents the number of video blocks (chronological segments), and the \textbf{y-axis} denotes the Aesthetic Score.}
    \label{fig:Aesth}
\end{figure}
\begin{figure}[htp]
    \centering
    \includegraphics[width=\linewidth]{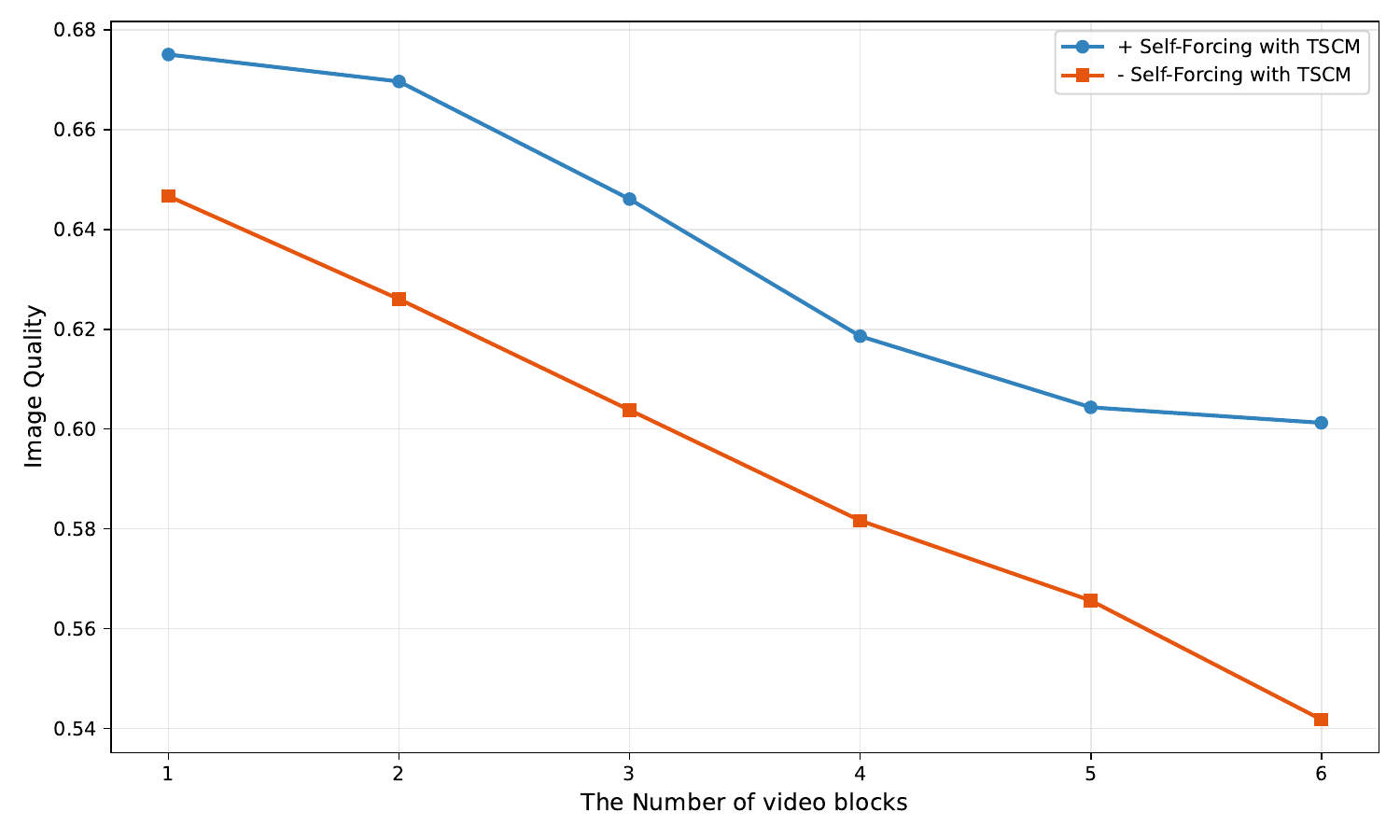}
    \caption{Image Quality Dynamics in Long-video Generation. The \textbf{x-axis} corresponds to the number of video blocks, and the \textbf{y-axis} shows the Image Quality score.}
    \label{fig:Image}
\end{figure}
\subsubsection{Validation of Long-video Generation Performance}
To evaluate the long-video generation capability of our model, we conducted text-to-video generation tests, which present greater challenges than image-to-video tasks. Since the original Yume-Bench test data tends to encounter limitations in long-video generation, we relabeled the test samples using InternVL3-78B to obtain improved captions for text-to-video generation and estimated forward movement controls. We compared models trained with and without Self-Forcing with TSCM, using 4 and 20 sampling steps respectively. For each 30-second generated video, we extracted 6 consecutive 5-second clips and computed both aesthetic quality and image quality scores for each segment, then analyzed how these metrics evolved over time. As shown in Figure~\ref{fig:Image}, the model trained with Self-Forcing and TSCM demonstrates more stable aesthetic scores from the 4th to 6th video segments, achieving a final aesthetic score of 0.523 in the 6th segment - notably higher than the 0.442 achieved by the model without Self-Forcing and TSCM. Similarly, Figure~\ref{fig:Aesth} reveals that the image quality scores of the Self-Forcing with TSCM model remain more consistent between the 5th and 6th segments, with the 6th segment attaining an image quality score of 0.601 compared to 0.542 for the baseline model.

\subsection{Ablation study}
\subsubsection{Verification of TSCM}
\begin{figure}[t!]
    \centering
    \includegraphics[width=\linewidth]{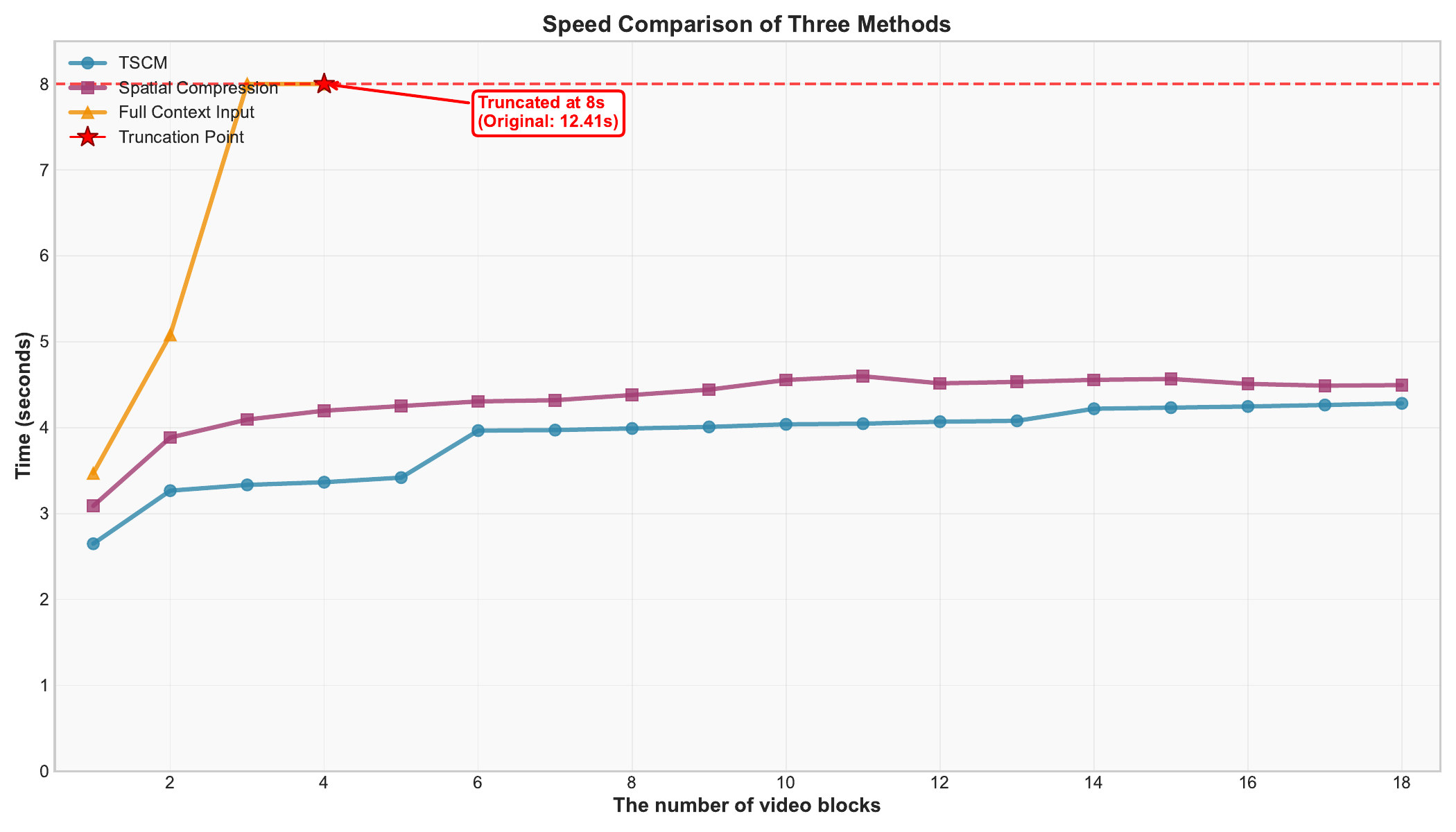}
    \caption{Speed Comparison of TSCM, Spatial Compression and Full Context Input. The test resolution is 704$\times$1280. The \textbf{x-axis} indicates the number of video blocks (increasing context length), and the \textbf{y-axis} represents the inference time in seconds.}
    \label{fig:speed}
\end{figure}

We established \method model with TSCM removed as the baseline. To validate the effectiveness of TSCM, we trained a new model under identical training configurations by removing TSCM and incorporating the spatial compression module from ~\cite{mao2025yume}. As shown in Table~\ref{tab:model_comparison}, our method demonstrates significant improvement in the key metric of Instruction Following. This is likely because TSCM reduces the influence of motion directions inherent in historical frames on predicted frames, while other metrics show minimal differences. We also compared the variation in autoregressive inference time across different methods as the number of video blocks increases, as illustrated in Figure~\ref{fig:speed}. The results show that the method using TSCM maintains the most stable change in autoregressive inference time with increasing video blocks, reducing fluctuations caused by expanding context. When the number of video blocks exceeds 8, the inference time per step remains constant. In contrast, the full-context input method exhibits the slowest performance, showing the largest gap compared to other methods at the 3rd inference step.

\begin{table}[t!]
    \renewcommand\arraystretch{1.1}
    \setlength{\tabcolsep}{1.6pt}
  \caption{Validation of distillation method effectiveness. IF: Instruction Following, SC: Subject Consistency, BC: Background Consistency, MS: Motion Smoothness, AQ: Aesthetic Quality, IQ: Imaging Quality.}
  \vspace{-3mm}
  \label{tab:model_comparison}
  \centering
  \begin{tabular}{l|c|c|c|c|c|c}
    \thickhline
    Model
    & IF$\uparrow$ & SC$\uparrow$ & BC$\uparrow$ & MS$\uparrow$ & AQ$\uparrow$ & IQ$\uparrow$ \\
    \hline
    TSCM&\textbf{0.836}&{0.932} &\textbf{0.945} &\textbf{0.985} &\textbf{0.506} &0.728  \\
    Spatial Compression&{0.767}&\textbf{0.935} &\textbf{0.945} &{0.973} &{0.504} &\textbf{0.733} \\
    \thickhline
  \end{tabular}
\end{table}


\subsection{Qualitative Results}
\begin{figure}[t]
    \centering
    \includegraphics[width=\linewidth]{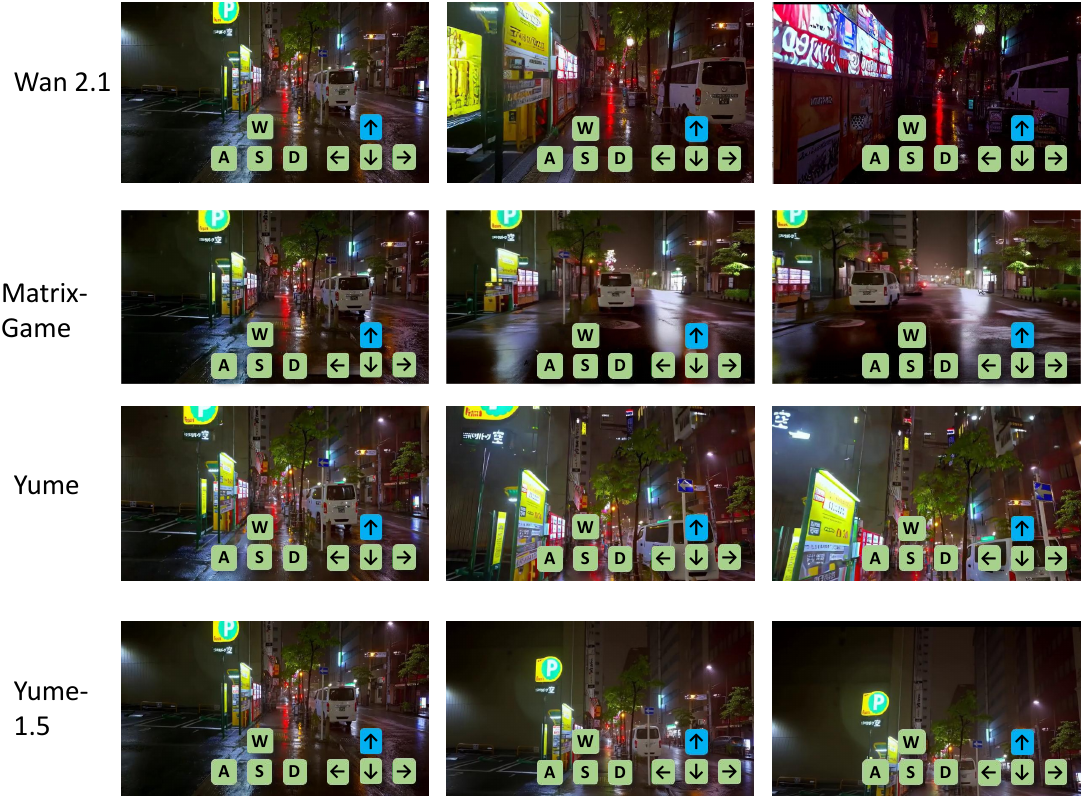} 
    \vspace{-2mm}
    \caption{Qualitative generation results. All tests were conducted at a resolution of 544$\times$960, with \method using 4 sampling steps while all other methods employed 50 sampling steps.}
    \label{fig:Qualitative}
\end{figure}

As shown in Figure~\ref{fig:Qualitative}, our method demonstrates effective camera control while achieving superior generation quality. Other visual results are included in the supplementary materials.

\section{Conclusion}
\label{sec:conclusion}

In this paper, we present \method, an interactive world generation model that enables infinite video generation from a single input image through autoregressive synthesis while supporting intuitive keyboard-based camera control. Our framework addresses three fundamental challenges in dynamic world generation: limited generalizability across domains, high computational latency, and insufficient text-based control capabilities. 

The key innovations of \method include: (1) a joint temporal-spatial-channel modeling approach that enables efficient long video generation while maintaining temporal coherence; (2) an acceleration method  that mitigates error accumulation during inference; and (3) text-controlled world event generation capability achieved through careful architectural design and mixed-dataset training.

We will envision extending \method to support more sophisticated world interactions and broader application scenarios in virtual environments and simulation systems.

\section{Limitations} 
\method still exhibits certain generation artifacts, such as vehicles moving backwards and characters walking in reverse. Performance tends to degrade in scenarios with extremely high crowd density. While increasing the resolution from 540p to 720p provides some mitigation, these issues persist to some extent. We attribute these limitations to the constrained capacity of the 5B parameter model; however, scaling to larger models would lead to prohibitively high generation latency. Inspired by Wan2.2, we consider exploring Mixture-of-Experts (MoE) architectures as a promising direction to achieve both larger parameter counts and reduced inference latency.